\pdfoutput=1
\documentclass[11pt]{article}
\providecommand{\paperoption}{preprint}
\edef\loadacl{\noexpand\usepackage[\paperoption]{acl}}\loadacl
\newif\ifanon
\def\tmp{review}\ifx\paperoption\tmp\anontrue\fi

\usepackage{times}
\usepackage{latexsym}
\usepackage[T1]{fontenc}
\usepackage[utf8]{inputenc}
\usepackage{microtype}
\IfFileExists{inconsolata.sty}{\usepackage{inconsolata}}{}  
\usepackage{graphicx}
\usepackage{float}
\usepackage{amsmath,amssymb,amsthm}
\usepackage{booktabs}
\usepackage{xcolor}
\usepackage{tikz}
\usetikzlibrary{arrows.meta,positioning,fit,calc,backgrounds,shapes.geometric}
\usepackage[scaled=0.90]{helvet}  
\usepackage{enumitem}
\usepackage{orcidlink}
\usepackage[capitalise,noabbrev]{cleveref}
\setlist{nosep,leftmargin=*}

\newtheorem{proposition}{Proposition}
\newcommand{\vstop}{\mathcal{V}_{\mathrm{stop}}}
\newcommand{\sys}{MinimalLM}
\definecolor{reduce}{HTML}{EB6834}
\definecolor{guard}{HTML}{1BAF7A}

\newcommand{\cmark}{\checkmark}

\newcommand{\nBaseCut}{5.7\%}

\newcommand{\nBaseJInc}{14.7\%}
\newcommand{\nBaseJMiss}{12.6\%}
\newcommand{\nBaseJScore}{8.42}

\newcommand{\nBatchSaveMax}{55\%}
\newcommand{\nBatchSaveMin}{5\%}
\newcommand{\nBudget}{200}

\newcommand{\nCkProposedTrimmedAtCap}{5}

\newcommand{\nGateCut}{0.3\%}

\newcommand{\nGateJInc}{21.7\%}

\newcommand{\nGateJScore}{8.22}

\newcommand{\nGateTRR}{19.8\%}

\newcommand{\nInstrCut}{0.6\%}

\newcommand{\nInstrJInc}{22.0\%}

\newcommand{\nInstrJScore}{8.07}

\newcommand{\nInstrTRR}{45.3\%}

\newcommand{\nInstrTRRMis}{17.5\%}
\newcommand{\nInstrTRRQwe}{68.4\%}

\newcommand{\nMinCut}{59.3\%}

\newcommand{\nMinInstrCut}{17.3\%}

\newcommand{\nMinInstrJInc}{31.6\%}

\newcommand{\nMinInstrJScore}{7.83}

\newcommand{\nMinInstrTRR}{50.8\%}

\newcommand{\nMinJComplete}{45\%}

\newcommand{\nMinJInc}{55.2\%}

\newcommand{\nMinJScore}{7.51}

\newcommand{\nMinTRR}{32.6\%}

\newcommand{\nOverheadMax}{0.8\%}
\newcommand{\nPropCov}{37\%}
\newcommand{\nPropCut}{0.0\%}

\newcommand{\nPropJComplete}{77\%}

\newcommand{\nPropJInc}{23.4\%}
\newcommand{\nPropJMiss}{22.9\%}
\newcommand{\nPropJScore}{8.04}
\newcommand{\nPropN}{990}
\newcommand{\nPropTRR}{47.6\%}
\newcommand{\nPropTRRLla}{47.5\%}
\newcommand{\nPropTRRMis}{23.1\%}
\newcommand{\nPropTRRQwe}{68.8\%}
\newcommand{\nPropTRRhi}{49.5\%}
\newcommand{\nPropTRRlo}{45.7\%}

\newcommand{\nTrAtPropComplete}{11\%}
\newcommand{\nTrAtPropCov}{58\%}

\newcommand{\nTrAtPropScore}{6.46}
\newcommand{\nTrTrimAtMinComplete}{57\%}

\newcommand{\nTrTrimAtMinScore}{7.36}
\newcommand{\nTrTrimAtPropComplete}{40\%}

\newcommand{\nTrTrimAtPropScore}{6.82}

\title{Say Less, Mean Better:\\Training-Free Verbosity Control for Language Models}

\ifanon
\author{Anonymous COLING 2027 Industry Track submission}
\else
\author{Mahmoud Alyosify\,\orcidlink{0009-0005-1336-2726}, Mirna Imbabi\,\orcidlink{0009-0000-6699-3170}, Eman Elkhamisi\,\orcidlink{0009-0002-6077-3593}, Rahatara Ferdousi\,\orcidlink{0000-0003-1143-2370} \\
  School of Computing, Queen's University, Kingston, Ontario, Canada \\
  \texttt{\{mahmoud.alyosify, mirna.imbabi, 25xbvh, rahatara.ferdousi\}@queensu.ca}}
\fi

\begin{document}
\maketitle

\begin{abstract}
Large language models (LLMs) often generate responses longer than necessary, increasing inference latency and computational cost. While prompting can encourage conciseness, controlling output length without compromising response completeness remains challenging. Aggressive early stopping can reduce generation costs but risks terminating responses prematurely. We introduce \textbf{MinimalLM}, a training-free framework for reliable verbosity control that combines uncertainty-aware, length-dependent stop-token adjustment with structural completeness guardrails. MinimalLM separates the decision of when to stop from whether stopping is structurally appropriate, coordinating response-level instructions, decoding-time intervention, and completion checks without modifying model parameters or requiring auxiliary models. We evaluate MinimalLM across three open-source LLMs and 330 questions, producing 5,940 responses across baseline and ablation configurations. MinimalLM reduces generated tokens by \textbf{47.6\%}, accelerates individual responses by \textbf{19--68\%}, and eliminates structurally truncated outputs in the primary evaluation. Despite substantial token savings, it achieves an average quality score of \textbf{8.04/10}, compared with \textbf{8.42/10} for unmodified models and \textbf{6.46/10} for length-matched truncation. Evaluations across diverse subject domains (such as healthcare, programming, law, and science) demonstrate robust efficiency gains, while ablations establish the complementary roles of decoding control and completeness constraints. Our findings highlight that efficient generation requires more than producing fewer tokens: it requires controlling when generation ends while preserving response integrity. MinimalLM offers a lightweight, training-free approach toward this goal.\looseness=-1 
\end{abstract}

\section{Introduction}

Large language models (LLMs) often generate responses longer than necessary, increasing inference latency and computational cost. Since autoregressive generation produces one token at a time, longer responses require additional decoding steps. This challenge is amplified by instruction tuning and preference optimization, which can favor verbose responses \cite{singhal2023long,saito2023verbosity}. Reducing unnecessary generation therefore offers a direct opportunity to improve efficiency. However, shorter responses are only useful when they preserve the information needed to address the user's request.\looseness=-1 

Existing inference optimization methods primarily focus on reducing the computational cost of generation. Speculative decoding accelerates token generation \cite{leviathan2023speculative}, while prompt-token pruning and sparse computation reduce processing overhead \cite{fu2024lazyllm,long2025sliminfer,federici2025efficient}. These approaches do not directly control the number of tokens in the final response. Other methods, such as TokenSkip \cite{xia2025tokenskip}, reduce generated tokens through model adaptation, requiring additional training. Training-free alternatives offer a lightweight solution, but introduce a fundamental challenge: controlling when generation should stop without prematurely terminating the response.

This challenge exposes an important distinction between \emph{reducing output length} and \emph{controlling verbosity reliably}. Length-constrained prompting can encourage concise responses, but provides limited control over the final token budget. Conversely, decoding-time stopping can enforce earlier termination without ensuring that the response ends at an appropriate boundary. A response may therefore achieve substantial token savings while remaining structurally incomplete. Reliable verbosity control requires coordinating the decision to stop with constraints on where stopping is permitted.

In this paper, we introduce \textbf{\sys{}}, a training-free framework for reliable verbosity control in autoregressive language models. The central idea is to separate \emph{when stopping becomes desirable} from \emph{whether stopping is structurally appropriate}. \sys{} combines a length-dependent, uncertainty-aware stop-token adjustment with a completeness gate that restricts termination to valid response boundaries. A lightweight instruction encourages the model to plan concise, complete responses, while a final check handles outputs that reach the generation limit. The framework operates at inference time without modifying model parameters or requiring auxiliary models.

Our investigation also reveals why decoding-time control must be evaluated beyond token reduction and text similarity. An initial stopping mechanism reduced generated tokens substantially, yet frequently terminated responses mid-sentence. Such failures can remain difficult to identify through similarity-based evaluation alone. By incorporating structural guardrails, \sys{} addresses this failure mode while retaining substantial efficiency gains. Importantly, we distinguish structural completeness from information coverage: ending a response at a valid boundary does not necessarily guarantee that every requested detail has been addressed.

We evaluate \sys{} across three open-source language models and 330 questions, generating 5,940 responses across baseline and ablation configurations. The framework reduces generated tokens by \nPropTRR{}, accelerates individual responses by 19--68\%, and produces no structurally truncated outputs in the primary evaluation. We further examine response quality, comparisons with length-matched truncation, batched serving, and performance across diverse subject domains.

Our contributions are threefold:
\begin{itemize}
    \item \textbf{Training-free verbosity control.}
    We introduce \sys{}, an inference-time framework that combines uncertainty-aware stopping with structural completeness constraints, without model retraining or auxiliary networks (\S\ref{sec:system}).

    \item \textbf{Reliability-aware evaluation.}
    We investigate the trade-off between output length, response quality, and structural completeness through 5,940 responses across three language models, including ablations and comparisons with direct truncation (\S\ref{sec:setup}--\S\ref{sec:results}).

    \item \textbf{Practical insights into efficient generation.}
    We analyze inference latency, batched serving, and domain-specific performance across diverse topics, while identifying limitations of prompt-only length control, similarity-based evaluation, and fixed response budgets (\S\ref{sec:lessons}).
    
\end{itemize}

\section{Related Work}
\label{sec:related}

\Cref{tab:related} places the system among the methods a service can use to cut generation cost. Speculative decoding \cite{leviathan2023speculative} makes decoding 2--3$\times$ faster with identical outputs; LazyLLM \cite{fu2024lazyllm} and SlimInfer \cite{long2025sliminfer} prune prompt tokens for long contexts; dynamic input pruning \cite{federici2025efficient} skips computation inside the model. All make each token cheaper and leave the number of output tokens unchanged, so they combine with ours. TokenSkip \cite{xia2025tokenskip} removes 40\% of reasoning tokens with LoRA fine-tuning. Asking the model to be concise is free but loose and model-dependent (\S\ref{sec:results}); a token cap or a stop-token boost \cite{holtzman2020curious} saves tokens but cuts answers wherever it stops. To our knowledge no training-free method both reduces generated tokens and guarantees that every answer ends complete; \sys{} does both, and is evaluated by counting incomplete answers rather than by similarity alone.

\begin{table}[t]
\centering
\small
\setlength{\tabcolsep}{2.2pt}
\begin{tabular}{@{}lccc@{}}
\toprule
Method & Fewer out. & Training- & No cut-off \\
 & tokens & free & answers \\
\midrule
Speculative decoding & -- & \checkmark & n/a \\
LazyLLM, SlimInfer & -- & \checkmark & n/a \\
Dynamic input pruning & -- & minimal & n/a \\
TokenSkip & \checkmark & -- & -- \\
``Be concise'' prompt & \checkmark & \checkmark & -- \\
Token cap, stop boost & \checkmark & \checkmark & -- \\
\textbf{\sys} & \checkmark & \checkmark & \checkmark \\
\bottomrule
\end{tabular}
\caption{Ways to cut generation cost. Only methods that reduce output tokens can cut answers off.}
\label{tab:related}
\end{table}

\section{\sys{}}
\label{sec:system}

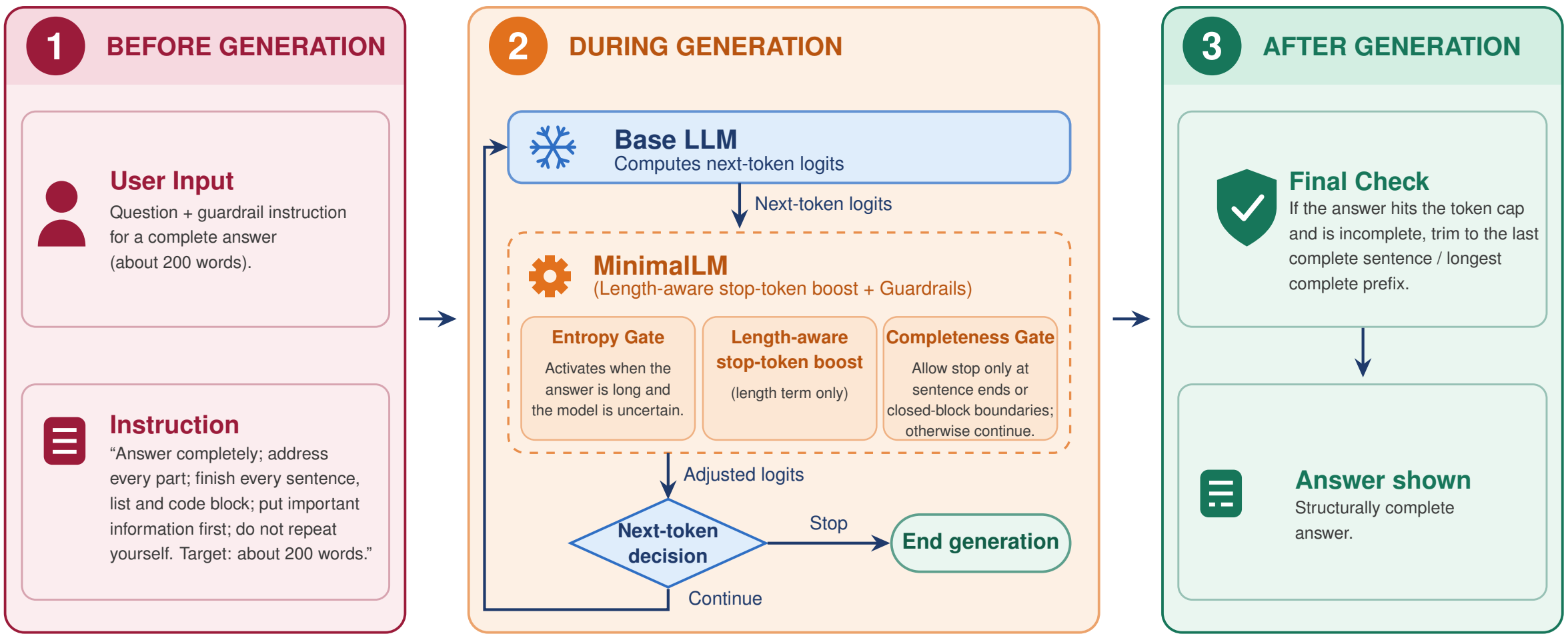
\begin{figure*}[t]
\centering
\resizebox{\textwidth}{!}{
\definecolor{archred}{HTML}{9B1B3A}\definecolor{archredpale}{HTML}{FBEDEF}\definecolor{archredband}{HTML}{F6D9E0}
\definecolor{archorange}{HTML}{E2711D}\definecolor{archorangetext}{HTML}{B85410}
\definecolor{archorangepale}{HTML}{FDF0E4}\definecolor{archorangecard}{HTML}{FBDDC2}
\definecolor{archgreen}{HTML}{17765A}\definecolor{archgreenpale}{HTML}{EAF6F0}\definecolor{archgreenband}{HTML}{D3EEE2}
\definecolor{archblue}{HTML}{2F6FC4}\definecolor{archbluepale}{HTML}{DFEDFB}\definecolor{archnavy}{HTML}{1F3A6B}
\definecolor{archtext}{HTML}{3A3A3A}
\begin{tikzpicture}[x=0.0125cm, y=-0.0125cm,
    font=\sffamily,
    >={Stealth[length=2.4mm, width=2.1mm]},
    arr/.style={->, line width=1.0pt, draw=archnavy},
    num/.style={circle, inner sep=0pt, minimum size=0.74cm, text=white, font=\sffamily\bfseries\fontsize{15}{15}\selectfont},
    stage/.style={anchor=west, font=\sffamily\bfseries\fontsize{10.1}{12}\selectfont},
    ctitle/.style={anchor=west, inner sep=0pt, font=\sffamily\bfseries\fontsize{10}{12}\selectfont},
    body/.style={anchor=north west, inner sep=0pt, align=left, text=archtext, font=\sffamily\fontsize{6.8}{9.0}\selectfont},
    lab/.style={inner sep=1pt, font=\sffamily\fontsize{7.4}{9}\selectfont, text=archnavy},
    card/.style={rounded corners=5pt, line width=0.7pt},
]
\foreach \x/\c/\p/\b in {15/archred/archredpale/archredband, 1181/archgreen/archgreenpale/archgreenband} {
  \fill[\p] [rounded corners=7pt] (\x,27) rectangle ++(403,630);
  \begin{scope}
    \clip[rounded corners=7pt] (\x,27) rectangle ++(403,630);
    \fill[\b] (\x,27) rectangle ++(403,78);
  \end{scope}
  \draw[\c, line width=0.9pt, rounded corners=7pt] (\x,27) rectangle ++(403,630);
}
\fill[archorangepale, draw=archorange!65, line width=0.9pt, rounded corners=7pt] (482,27) rectangle (1117,657);
\node[num, fill=archred] at (66,66) {1};
\node[stage, text=archred] at (106,66) {BEFORE GENERATION};
\node[num, fill=archorange] at (532,66) {2};
\node[stage, text=archorangetext] at (572,66) {DURING GENERATION};
\node[num, fill=archgreen] at (1231,66) {3};
\node[stage, text=archgreen] at (1271,66) {AFTER GENERATION};
\draw[arr] (432,341) -- (470,341);
\draw[arr] (1131,341) -- (1169,341);

\draw[card, archred!40, fill=archredpale] (32,132) rectangle (402,349);
\fill[archred] (72,218) circle (16);
\fill[archred] (48,268) .. controls (48,232) and (96,232) .. (96,268) -- cycle;
\node[ctitle, text=archred] at (120,204) {User Input};
\node[body] at (120,226) {Question + guardrail instruction\\for a complete answer\\(about 200 words).};
\draw[card, archred!40, fill=archredpale] (32,407) rectangle (402,624);
\fill[archred, rounded corners=3pt] (54,440) rectangle (96,488);
\foreach \y in {453,464,475} \draw[white, line width=1.4pt] (63,\y) -- (87,\y);
\node[ctitle, text=archred] at (120,447) {Instruction};
\node[body] at (120,469) {``Answer completely; address\\every part; finish every sentence,\\list and code block; put important\\information first; do not repeat\\yourself. Target: about 200 words.''};

\draw[card, archblue!75, fill=archbluepale, line width=0.9pt] (522,132) rectangle (1088,204);
\begin{scope}[shift={(567,168)}, archblue, line width=1.1pt, line cap=round]
  \foreach \a in {0,60,120} {
    \draw[rotate=\a] (-22,0) -- (22,0);
    \foreach \s in {1,-1} \draw[rotate=\a] (\s*14,0) -- (\s*20,6) (\s*14,0) -- (\s*20,-6);
  }
\end{scope}
\node[ctitle, text=archnavy, font=\sffamily\bfseries\fontsize{10.4}{12}\selectfont] at (628,160) {Base LLM};
\node[anchor=west, inner sep=0pt, text=archnavy, font=\sffamily\fontsize{7.6}{9}\selectfont] at (628,186) {Computes next-token logits};
\draw[arr] (755,204) -- (755,250) node[lab, midway, right=4pt] {Next-token logits};
\draw[archorange!80, fill=archorangepale, line width=0.8pt, dashed, rounded corners=6pt] (522,254) rectangle (1088,476);
\begin{scope}[shift={(564,298)}]
  \foreach \a in {0,45,...,315} \fill[archorange, rotate=\a] (-5,-12) rectangle (5,-21);
  \fill[archorange] (0,0) circle (15.5);
  \fill[archorangepale] (0,0) circle (6.5);
\end{scope}
\node[ctitle, text=archorangetext, font=\sffamily\bfseries\fontsize{10.4}{12}\selectfont] at (607,287) {MinimalLM};
\node[anchor=west, inner sep=0pt, text=archorangetext, font=\sffamily\fontsize{7.5}{9}\selectfont] at (607,312) {(Length-aware stop-token boost + Guardrails)};
\foreach \x/\t/\b in {
    535/{Entropy Gate}/{Activates when the\\answer is long and\\the model is uncertain.},
    718/{Length-aware\\stop-token boost}/{(length term only)},
    900/{Completeness Gate}/{Allow stop only at\\sentence ends or\\closed-block boundaries;\\otherwise continue.}} {
  \draw[card, archorange!55, fill=archorangecard!70, line width=0.6pt] (\x,339) rectangle ++(175,124);
  \node[anchor=north, align=center, inner sep=0pt, text=archorangetext, font=\sffamily\bfseries\fontsize{7.2}{8.8}\selectfont]
    (t) at (\x+87.5,352) {\t};
  \node[anchor=north, align=center, inner sep=0pt, text=archtext, font=\sffamily\fontsize{6.0}{7.5}\selectfont]
    at ([yshift=-0.18cm]t.south) {\b};
}
\draw[arr] (683,476) -- (683,523) node[lab, midway, right=4pt] {Adjusted logits};
\node[diamond, aspect=2.2, draw=archblue, fill=archbluepale, line width=0.9pt, inner sep=0.5pt,
      align=center, text=archnavy, font=\sffamily\bfseries\fontsize{7.8}{9}\selectfont] (dec) at (683,567) {Next-token\\decision};
\node[rounded corners=0.36cm, draw=archgreen!70, fill=archgreenpale, line width=0.9pt, minimum width=2.25cm,
      minimum height=0.72cm, text=archgreen!80!black, font=\sffamily\bfseries\fontsize{8.6}{10}\selectfont] (end) at (998,567) {End generation};
\draw[arr] (dec.east) -- (end.west) node[lab, midway, above=2pt] {Stop};
\draw[arr] (dec.south) -- (683,634) -- (498,634) -- (498,168) -- (522,168);
\node[lab, anchor=west] at (700,622) {Continue};

\draw[card, archgreen!45, fill=archgreenpale] (1197,132) rectangle (1568,349);
\begin{scope}[shift={(1266,228)}]
  \fill[archgreen] (0,-38) -- (30,-27) -- (30,0) .. controls (30,20) and (14,32) .. (0,40)
    .. controls (-14,32) and (-30,20) .. (-30,0) -- (-30,-27) -- cycle;
  \draw[white, line width=1.8pt, line cap=round, line join=round] (-13,2) -- (-3,12) -- (15,-10);
\end{scope}
\node[ctitle, text=archgreen] at (1308,202) {Final Check};
\node[body] at (1308,222) {If the answer hits the token cap\\and is incomplete, trim to the last\\complete sentence / longest\\complete prefix.};
\draw[arr] (1383,350) -- (1383,400);
\draw[card, archgreen!45, fill=archgreenpale] (1197,407) rectangle (1568,624);
\fill[archgreen, rounded corners=3pt] (1219,493) rectangle (1261,541);
\foreach \y in {506,517} \draw[white, line width=1.4pt] (1228,\y) -- (1252,\y);
\draw[white, line width=1.4pt] (1228,528) -- (1234,528); \draw[white, line width=1.4pt] (1240,528) -- (1252,528);
\node[ctitle, text=archgreen] at (1314,503) {Answer shown};
\node[body] at (1314,524) {Structurally complete\\answer.};
\end{tikzpicture}}
\caption{\sys{} in three stages: an instruction before generation; inside the decoding loop, the stop-token boost with its entropy gate and the completeness gate; and a final check after generation. The base LLM is kept unchanged.}
\label{fig:system}
\end{figure*}

As illustrated in \Cref{fig:system}, \sys{} operates in three stages: response guidance before generation, controlled stopping during decoding, and structural validation after generation. The framework combines lightweight inference-time mechanisms to reduce unnecessary generation while preventing premature termination at incomplete~response~boundaries.

\paragraph{Stop, not skip.}
Autoregressive language models generate each token conditioned on the preceding sequence. Consequently, a decoder cannot skip a redundant passage and resume generation as though that passage had never been produced. Removing already generated text does not recover its computational cost. At each decoding step, the model can continue generation or terminate through a stop token, while modifying other token choices may alter the subsequent response.

Our preliminary investigation found that a semantic redundancy signal changed only 12 of 24,021 greedy decoding decisions. Moreover, stopping at redundant content often discarded useful information appearing later in the response. These observations motivate \sys{} to focus on controlled early termination rather than token-level content removal. The framework combines response-level length guidance with decoding-time stopping and structural safeguards.

\paragraph{Length- and uncertainty-aware stopping.}
At decoding step $t$, a logits processor adjusts only the scores $z_t$ associated with stop tokens $\vstop$:
\begin{equation}
\tilde z_t(s) = z_t(s) +
b_t\,g(H_t)\,w_L\,\ell(t)\,m_t,
\quad s \in \vstop,
\label{eq:proc}
\end{equation}
where
\begin{equation}
\ell(t) =
\mathrm{clip}
\left(
\frac{t-t_{\min}}{t_{\max}-t_{\min}},
0,1
\right)^{\gamma}
\end{equation}
increases the stopping preference as generation progresses. The uncertainty-dependent factor $g(H_t)$ is determined by the entropy of the next-token distribution, while
\begin{equation}
m_t =
\max_v z_t(v) -
\max_{s\in\vstop} z_t(s)
\end{equation}
represents the logit gap between the highest-scoring token and the highest-scoring stop token.

The structural gate $b_t$ controls whether the adjustment is permitted. Without this gate ($b_t=1$), the adjusted stop score reaches or exceeds the highest original token score when $g(H_t)w_L\ell(t)\geq 1$, assuming a positive logit gap.

We use $w_L=1.1$, $\gamma=1.5$, and an entropy band of $(0.05,0.6)$ nats. The model-specific thresholds $(t_{\min},t_{\max})$ are selected near the 25th and 50th percentiles of response lengths observed on development questions. Under these settings, the stopping intervention becomes effective only after approximately 380 generated tokens.

Importantly, the processor modifies only stop-token logits. Under greedy decoding, when the original generation has not already terminated, the resulting output is therefore a prefix of the unmodified response. The mechanism controls termination rather than rewriting generated content.

\paragraph{Structural completeness gate.}
The central design principle of \sys{} is to distinguish \emph{when stopping becomes desirable} from \emph{whether stopping is structurally appropriate}. The stopping adjustment is enabled only when the generated text reaches a recognized completion boundary:
\begin{equation}
b_t =
\begin{cases}
1, & \text{if a valid completion boundary is detected},\\
0, & \text{otherwise}.
\end{cases}
\end{equation}

Valid boundaries include sentence-ending punctuation outside abbreviations and list markers, closed code fences, and balanced brackets. When no valid boundary is detected, the gate suppresses the stopping adjustment and generation continues. Boundary validation is performed only when the stopping mechanism would otherwise trigger termination, limiting additional decoding overhead.

This gate prevents the stopping intervention from terminating responses at recognized incomplete boundaries. It does not, however, establish whether all information requested by the user has been covered.\looseness=-1 

\paragraph{Response-level instruction.}
Before decoding, a system instruction asks the model to provide a complete response within approximately \nBudget{} words, address the requested information, and finish sentences and structured content (Appendix~\ref{app:prompt}). This encourages the model to organize its response within a soft length budget.

The instruction and decoding mechanism serve complementary purposes. The instruction encourages concise generation, while the stop-token adjustment provides additional termination pressure when the model exceeds its expected response length. The structural gate restricts this intervention to recognized completion boundaries.

\paragraph{Final completion check.}
As a final safeguard, responses reaching the maximum generation limit are checked for structural completeness. If the output ends mid-sentence, it is trimmed to the last recognized complete sentence. This post-processing step requires no additional model inference.

Together, the completeness gate and final check provide the following structural property.

\begin{proposition}
Under the implemented boundary-detection rules, the stopping intervention and final completion check ensure that the displayed response ends at a recognized complete sentence or closed structural block, provided such a~boundary~exists.
\end{proposition}

This property concerns structural termination rather than semantic completeness. A response may end at a valid boundary while omitting information requested by the user.

\paragraph{Implementation and deployment.}
\sys{} requires a system instruction, a custom logits processor within the Hugging Face \texttt{generate} interface \cite{wolf2020transformers}, and a lightweight post-generation completion check:

\par\smallskip\noindent{\scriptsize\ttfamily
\hspace*{0.6em}proc = MinimalLMProcessor(stops, plen,\\
\hspace*{2.6em}t\_min=253, t\_max=405, tokenizer=tok)\\
\hspace*{0.6em}out = model.generate(**enc, max\_new\_tokens=1024,\\
\hspace*{2.6em}logits\_processor=[proc])\\
\hspace*{0.6em}text = final\_check(tok.decode(out[0, plen:]))}
\par\smallskip\noindent

The framework requires neither model retraining nor auxiliary networks and can be enabled at inference time for models exposing next-token logits.

Two parameters are particularly relevant for deployment: the instruction-level word budget, which influences the trade-off between conciseness and information coverage, and the stopping thresholds $(t_{\min},t_{\max})$, which are calibrated using typical model response lengths. The boundary-detection mechanism can also serve as a lightweight diagnostic for identifying structurally truncated responses under alternative length-control policies.

\section{Evaluation Setup}
\label{sec:setup}

\paragraph{Models and questions} Mistral-7B-Instruct-v0.2 \cite{jiang2023mistral}, Llama-3.1-8B-Instruct \cite{dubey2024llama3} and Qwen2.5-3B-Instruct \cite{qwen2024qwen25}, pinned revisions, fp16, greedy decoding, at most 1,024 new tokens, one NVIDIA A40. The 330 evaluation questions are the first turns of MT-Bench \cite{zheng2023judging} (80), and 150 AlpacaEval \cite{dubois2024length} and 100 Arena-Hard-Auto \cite{li2024crowdsourced} prompts, none used in development. To evaluate domain generalization across different topics, we further test questions spanning eight distinct subject categories: health, history, economics, programming, science, mathematics, writing, and law.

\paragraph{Systems} Six systems isolate each part: the baseline, the instruction only, \sys{} without guardrails (the boost alone), without the gate, without the instruction, and \sys{}~(5,940~answers).

\paragraph{Measures} Token reduction ratio $\mathrm{TRR}=1-\sum n_{\mathrm{sys}}/\sum n_{\mathrm{base}}$ over \emph{generated} tokens; time per answer; \emph{cut off} by a fixed rule (an answer ended by the boost or the cap mid-sentence, or inside an open structure); an LLM judge from another developer, Phi-4 \cite{abdin2024phi4}, blind to the system, which says whether the answer is cut off or misses requested parts and scores it 1--10; and content coverage of the baseline answer by NLI \cite{liu2019roberta}. A reading of 96 answers, blind to the automatic labels, checks the rule ($\kappa{=}1.00$) and the judge ($\kappa{=}0.97$ on cut-off answers, $0.26$ on missing parts, where it over-reports).

\section{Results}
\label{sec:results}

\begin{table}[t]
\centering
\small
\setlength{\tabcolsep}{2.6pt}
\begin{tabular}{@{}lcccc@{}}
\toprule
System & TRR & Cut off & Incompl. & Score \\
\midrule
Baseline & -- & \nBaseCut & \nBaseJInc & \nBaseJScore \\
Instruction only & \nInstrTRR & \nInstrCut & \nInstrJInc & \nInstrJScore \\
w/o guardrails & \nMinTRR & \nMinCut & \nMinJInc & \nMinJScore \\
w/o gate & \nMinInstrTRR & \nMinInstrCut & \nMinInstrJInc & \nMinInstrJScore \\
w/o instruction & \nGateTRR & \nGateCut & \nGateJInc & \nGateJScore \\
\textbf{\sys} & \textbf{\nPropTRR} & \textbf{\nPropCut} & \nPropJInc & \nPropJScore \\
\bottomrule
\end{tabular}
\caption{\sys{} and its ablations, with 990 answers
per configuration (330 questions $\times$ 3 models).
TRR denotes token reduction ratio; cut off indicates
structural truncation detected by the rule; incomplete
indicates cut-off or missing information identified
by the judge; score is the judge's rating (1--10).
The abbreviation w/o denotes removal of a component.}
\label{tab:main}
\end{table}

\paragraph{Main results.}
\Cref{tab:main} summarizes the performance of
\sys{} and its ablations. Across three models,
\sys{} reduces generated tokens by \nPropTRR{}
(95\% CI: \nPropTRRlo--\nPropTRRhi), with no
structurally truncated responses among
\nPropN{} evaluated outputs. The average quality
score remains competitive at \nPropJScore{},
compared with \nBaseJScore{} for the baseline.
Token savings result primarily from reduced
generation rather than post-processing; only
\nCkProposedTrimmedAtCap{} of \nPropN{} responses
require trimming at the generation limit.
Across individual models, token reductions are
\nPropTRRMis{} for Mistral-7B,
\nPropTRRLla{} for Llama-3.1-8B, and
\nPropTRRQwe{} for Qwen2.5-3B.
Output comparison for different topics are represented in Appendix~\ref{app:output-comparison}.

\begin{figure}[t]
\centering
\includegraphics[width=\columnwidth]{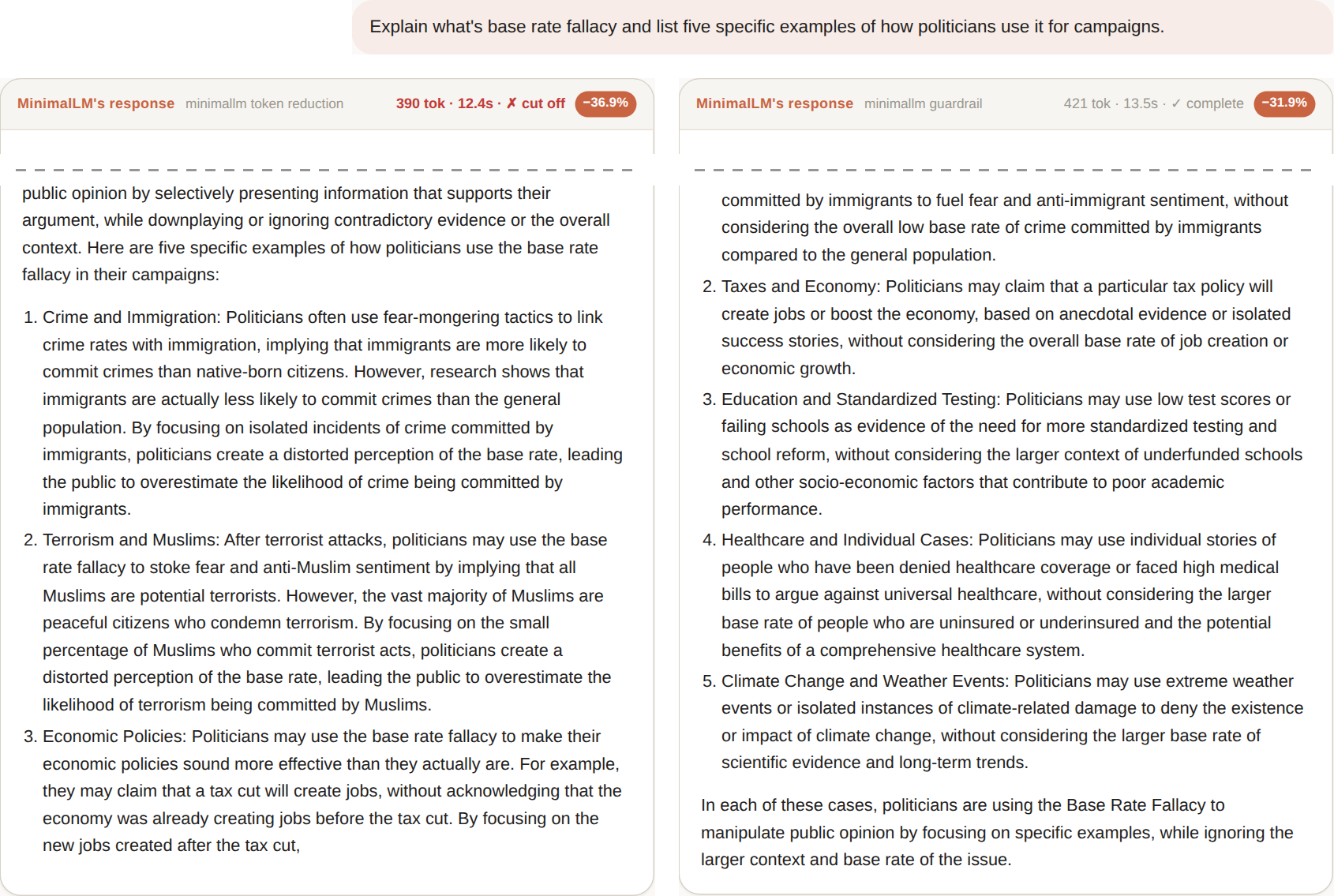}
\caption{Example responses from Mistral-7B to a
question requesting five examples, displayed in
our demonstration interface (intermediate text
omitted). Left: without guardrails, the stopping
mechanism terminates mid-sentence during the
third example (390 tokens). Right: \sys{}
provides all five examples and ends at a complete
sentence (421 tokens), compared with 618 tokens
for the baseline.}
\label{fig:example}
\end{figure}

\paragraph{Ablation analysis.}
The ablations in \Cref{tab:main} demonstrate
the complementary roles of the framework's
components. Without guardrails, the stopping
mechanism reduces tokens by \nMinTRR{} but
produces structurally truncated responses in
\nMinCut{} of cases. On MT-Bench, 82\% of
shortened responses terminate prematurely,
sometimes omitting later requested items
(\cref{fig:example}).

Instruction-only prompting achieves
\nInstrTRR{} token reduction, but its effectiveness
varies across models, from \nInstrTRRMis{}
on Mistral-7B to \nInstrTRRQwe{} on Qwen2.5-3B.
Removing the completeness gate results in
\nMinInstrCut{} truncated outputs, whereas
removing the instruction reduces token savings
to \nGateTRR{}. Together, these findings show
that prompting contributes substantially to
conciseness, while guarded decoding provides
additional control over response termination.
For Mistral-7B, the combined framework improves
token reduction from \nInstrTRRMis{} with
instruction-only prompting to \nPropTRRMis{}.

\begin{figure*}[t]
\centering
\includegraphics[width=\textwidth]{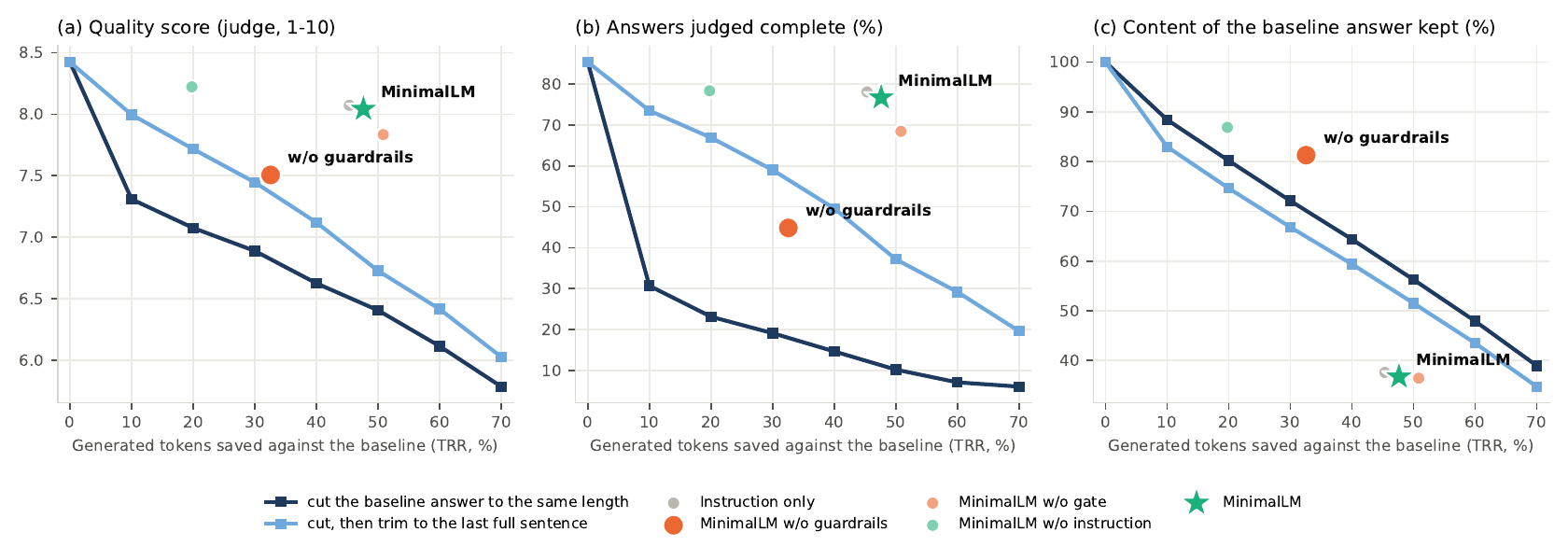}
\caption{Comparison against direct truncation
across 330 questions and three models.
Dark curves represent baseline responses
truncated to 90--30\% of their original length;
light curves represent the same responses
trimmed to the last complete sentence.
Points indicate the configurations in
\cref{tab:main}. At comparable token reductions,
\sys{} achieves higher judged quality and
completeness than direct truncation, although
it retains less content from the original
baseline response.}
\label{fig:trunc}\vspace{8pt}
\end{figure*}

\paragraph{Comparison with truncation.}
We compare \sys{} against direct truncation
of baseline responses at different token budgets,
with and without trimming to the last complete
sentence (\cref{fig:trunc}). At the token
reduction achieved by \sys{}, directly truncated
responses obtain a quality score of
\nTrAtPropScore{}, increasing to
\nTrTrimAtPropScore{} after sentence-level
trimming, compared with \nPropJScore{} for
\sys{}. The corresponding proportions judged
complete are \nTrAtPropComplete{},
\nTrTrimAtPropComplete{}, and
\nPropJComplete{}, respectively.

Without guardrails, the stopping mechanism
performs similarly to truncation, scoring
\nMinJScore{} against \nTrTrimAtMinScore{}
for sentence-trimmed truncation, with lower
judged completeness (\nMinJComplete{} vs.\
\nTrTrimAtMinComplete). These results indicate
that the improvement arises from coordinating
response generation with stopping constraints,
rather than early termination alone.
\sys{} also retains less content from the
original baseline response (\nPropCov{} vs.\
\nTrAtPropCov), reflecting differences between
generating a concise response and truncating
an existing longer one.

\begin{table}[t]
\centering
\small
\setlength{\tabcolsep}{3pt}
\begin{tabular}{@{}lcccc@{}}
\toprule
 & \multicolumn{2}{c}{ms / token}
 & \multicolumn{2}{c}{batched GPU time} \\
\cmidrule(lr){2-3}\cmidrule(lr){4-5}
Model & base & ours & base & ours \\
\midrule
Mistral-7B & 30.8 & 31.0 & 13.9 & 13.2 \\
Llama-3.1-8B & 32.2 & 32.5 & 13.8 & 10.3 \\
Qwen2.5-3B & 28.1 & 28.2 & 7.1 & 3.2 \\
 
\bottomrule
\end{tabular}
\caption{Inference and serving costs.
Milliseconds per generated token are measured
for individual requests using 16 MT-Bench
questions. Batched GPU time is reported in
minutes for 330 questions processed in
fixed batches.}
\label{tab:serving}
\end{table}

\paragraph{Inference latency and serving.}
Reducing generated tokens also improves
single-request latency. Across the evaluated
models, responses are 19--68\% faster,
including reductions from 10.1\,s to 8.2\,s
for Mistral-7B and from 10.8\,s to 3.5\,s
for Qwen2.5-3B. The additional decoding
overhead remains below \nOverheadMax{} per
token (\cref{tab:serving}).

Under fixed-batch serving, GPU time decreases
by \nBatchSaveMin--\nBatchSaveMax, indicating
that token reduction does not translate
directly into proportional throughput gains.
Because batch completion depends on the
longest-running response, variation in output
length can limit the realized savings.
Continuous batching \cite{kwon2023pagedattention}
may offer opportunities to better utilize
earlier completion, although this configuration
is not evaluated here.

\begin{table}[t]
\centering
\small
\setlength{\tabcolsep}{2.2pt}
\begin{tabular}{@{}lcccc@{}}
\toprule
Domain & Tokens & Change & Time (s) & Complete \\
\midrule
Health & 669 / 407 & $-$39\% & 21.1 / 12.9 & \cmark / \cmark \\
History & 645 / 405 & $-$37\% & 20.4 / 12.9 & \cmark / \cmark \\
Economics & 520 / 383 & $-$26\% & 16.4 / 12.2 & \cmark / \cmark \\
Programming & 784 / 300 & $-$62\% & 24.7 / 9.5 & \cmark / \cmark \\
Science & 629 / 330 & $-$48\% & 19.9 / 10.5 & \cmark / \cmark \\
Mathematics & 183 / 260 & +42\% & 5.8 / 8.3 & \cmark / \cmark \\
Writing & 374 / 288 & $-$23\% & 11.8 / 9.2 & \cmark / \cmark \\
Law & 429 / 286 & $-$33\% & 13.6 / 9.1 & \cmark / \cmark \\
 
\bottomrule
\end{tabular}
\caption{Domain performance on eight categories using Mistral-7B.
Values report baseline / \sys{} for generated
tokens, percentage change, latency in seconds,
and structural completeness (\cmark)
(see more examples in Appendix~\ref{app:output-comparison}).}
\label{tab:domain}
\end{table}

\paragraph{Category-wise domain evaluation.}
We further evaluate \sys{} across eight distinct
subject categories: health, history, economics,
programming, science, mathematics, writing, and law.
As shown in \Cref{tab:domain}, \sys{} consistently
reduces output length across seven of the eight domains,
yielding substantial token savings ranging from 23\% in
creative writing up to 62\% in programming (e.g., cutting
programming response tokens from 784 to 300 and latency
from 24.7\,s to 9.5\,s). Every response generated by
\sys{} terminates at a valid structural boundary with 100\%
completeness (\cmark). On mathematics, where the baseline
answer was already very brief (183 tokens), the target budget
leads to a token increase (+42\%), illustrating the
inherent trade-off of fixed-length budgeting on concise
prompts. Detailed side-by-side output comparisons for each
domain are presented in \Cref{app:output-comparison}.\looseness=-1 

\paragraph{Information coverage.}
The efficiency gains introduce a trade-off
in information coverage. The automatic judge
identifies missing requested information in
\nPropJMiss{} of \sys{} responses, compared
with \nBaseJMiss{} for the baseline.
However, agreement with manual assessments
is limited ($\kappa=0.26$), suggesting that
these estimates should be interpreted
cautiously. In a manually inspected sample,
none of 24 \sys{} responses was judged
incomplete, compared with 16 of 24 responses
generated without guardrails. These findings
motivate further investigation of response
coverage under constrained generation budgets.

\section{Discussion}
\label{sec:lessons}

Our findings highlight several considerations for reliable
verbosity control in autoregressive language models.

\begin{enumerate}[label=\textbf{\arabic*.}]

\item \textbf{Similarity does not guarantee completeness.}
The stop-token boost alone satisfied the predefined
BERTScore similarity margins, despite frequently
terminating responses mid-sentence. Since early
stopping preserves a prefix of the original output,
high textual similarity may conceal structural
incompleteness. This highlights the importance of
evaluating generated responses beyond similarity
metrics, using explicit completion checks and
manual inspection.

\item \textbf{Early stopping requires appropriate baselines.}
A direct truncation baseline provides an important
reference for evaluating output-length reduction.
Without structural guardrails, the stopping mechanism
offers limited improvement over truncation.
Combining response-level instructions with guarded
stopping yields higher response quality at comparable
token budgets, demonstrating the value of coordinating
generation and termination.

\item \textbf{Length control and structural completeness
are complementary.}
Prompt-based instructions encourage concise responses
but provide inconsistent length control across models.
Conversely, decoding-time stopping can reduce output
length while interrupting sentences or structured
content. \sys{} combines these mechanisms to achieve
substantial token savings without structurally
truncated responses in the primary evaluation.
However, structural completeness does not guarantee
that every requested detail is preserved.

\item \textbf{Token reduction does not directly translate
into proportional serving gains.}
Although shorter responses reduce individual inference
latency, improvements in batched serving depend on
scheduling and response-length variation. In static
batches, longer generations can limit the computational
benefits of shorter outputs. This distinction is
important when interpreting token savings as
system-level efficiency improvements.

\item \textbf{Verbosity reduction varies across domains.}
Domain-specific evaluation demonstrates substantial
token reductions in verbose areas such as programming
($-62\%$) and science ($-48\%$), where explanations
naturally invite verbosity. Conversely, in domains with
inherently concise responses (such as elementary mathematics),
fixed word budgets can prevent aggressive length reduction
or slightly extend concise answers. Tailoring stopping
criteria to domain requirements represents an important
avenue for further refinement.

\item \textbf{Fixed budgets introduce a coverage trade-off.}
A uniform target of approximately \nBudget{} words
can constrain responses to broad or multi-part
questions while unnecessarily extending some
short answers. This suggests that effective verbosity
control should account for the information requirements
of individual requests rather than relying exclusively
on a fixed response budget.\looseness=-1 

\end{enumerate}

\section{Conclusion}

We introduced \sys{}, a training-free framework for
controlling LLM response verbosity through
uncertainty-aware stopping and structural completeness
guardrails. Across three open-source models, the
framework reduces generated tokens by \nPropTRR{}
while maintaining competitive response quality and
avoiding structurally truncated outputs in the primary
evaluation. Additional experiments demonstrate
consistent token savings across diverse subject domains
and identify practical considerations for inference
latency and batched serving. Importantly, our findings
distinguish structural completeness from information
coverage, highlighting the limitations of fixed
response budgets. Overall, \sys{} demonstrates how
lightweight decoding-time control can improve
generation efficiency without model retraining,
while motivating future work on adaptive verbosity
control that better preserves requested information.

\section*{Limitations}

Our evaluation covers three open-source models
(3B--8B parameters) using greedy decoding on a
single GPU type. Larger models, alternative decoding
strategies, and hardware configurations remain
unexplored. Response quality was assessed primarily
through automatic evaluation, supplemented by
manual inspection. The domain experiments evaluated
single representative queries per category under a fixed
budget; broader topic-specific benchmarks and adaptive
response budgets could further refine performance across
varied user intent.

\section*{Ethical Considerations}

The models, datasets and judge are public; no personal data was collected. Fewer generated tokens also mean less energy per answer.

\bibliography{references}

\appendix

\section{The Instruction}
\label{app:prompt}
\begin{quote}\small
You are a helpful assistant. Always give a complete answer: address every part of the user's request, and finish every sentence, list and code block that you start. Never stop in the middle of a sentence. Put the most important information first and do not repeat yourself. Keep the whole answer within about 200 words, and plan it so that everything requested fits and the answer ends with a complete sentence.
\end{quote}

\section{Example}
\label{app:example}
Mistral-7B, ``Explain base rate fallacy and list five specific examples of how politicians use it for campaigns.'' The baseline writes 618 tokens. Without guardrails, \sys{} stops at 390 tokens, in the middle of the third example: ``\dots By focusing on the new jobs created after the tax cut,''. \sys{} writes 421 tokens with all five examples and a closing sentence.

\section{Reproducibility}
All prompts, answers, scores, judge verdicts and analysis scripts are released\ifanon{} at an anonymized repository for review\else{} at \url{https://github.com/MinimalLM/minimal-lm}\fi; one script recomputes every number of this paper from the stored answers. An interactive Google Colab notebook containing the core implementation code is available at \url{https://colab.research.google.com/drive/13m2_uiBbhu_6hGBDC7DP2wyFE756_PEz?usp=sharing}, allowing readers to run \sys{} with Qwen2.5-3B on free GPU resources.

\section{Output Comparison Across Domains}
\label{app:output-comparison}

The following screenshots show the MinimalLM interface comparing baseline and MinimalLM responses side by side across eight distinct subject domains (healthcare, history, economics, law, science, programming, creative writing, and mathematics). Each screenshot displays the baseline response on the left and the MinimalLM response on the right, illustrating the token reduction while maintaining response completeness.

\begin{figure*}[p]
\centering
\includegraphics[width=\textwidth,height=0.82\textheight,keepaspectratio]{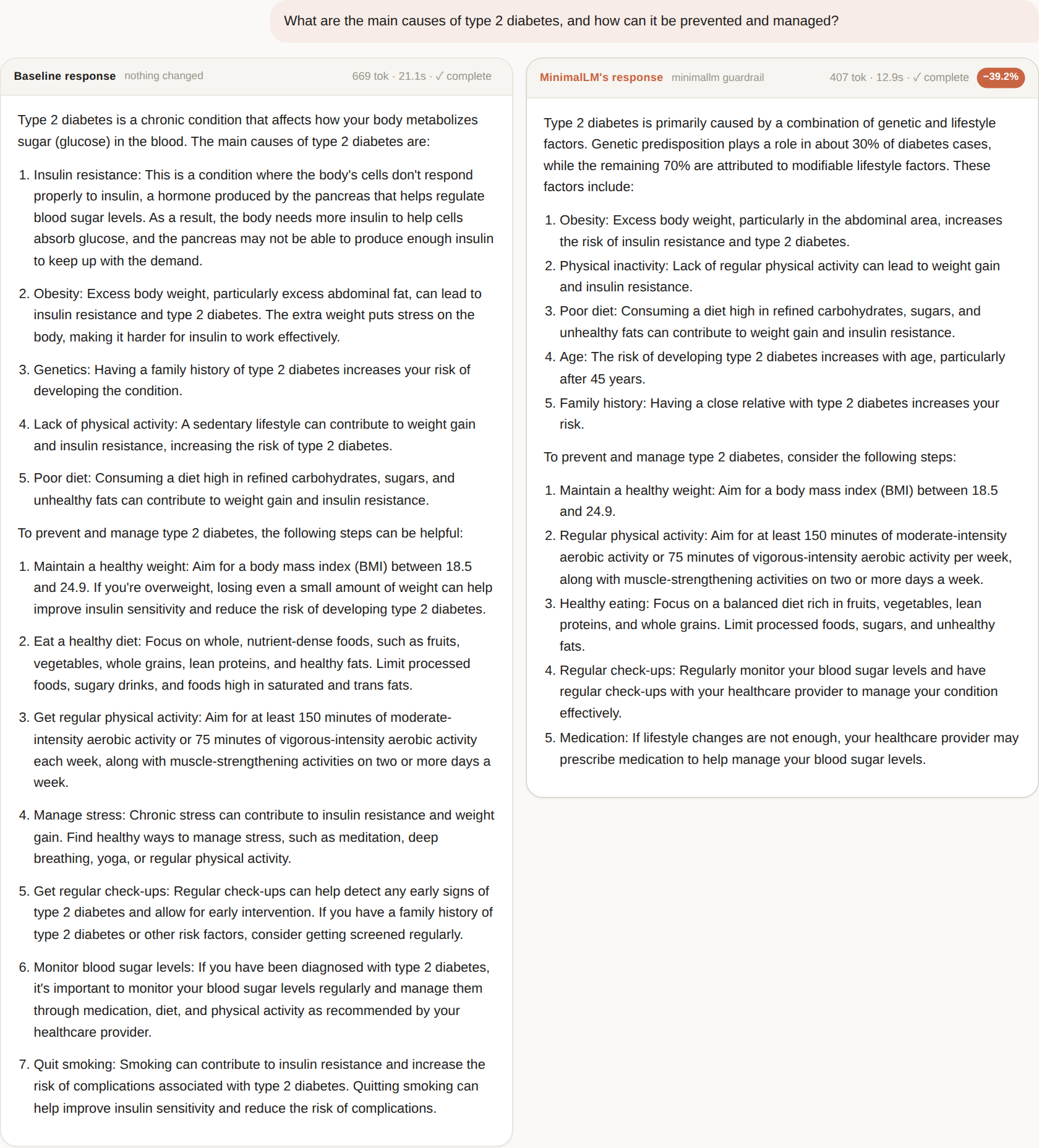}
\caption{Interface comparison on a health question (Mistral-7B). Left: baseline; right: MinimalLM.}
\label{fig:demo-health}
\end{figure*}

\begin{figure*}[p]
\centering
\includegraphics[width=\textwidth,height=0.82\textheight,keepaspectratio]{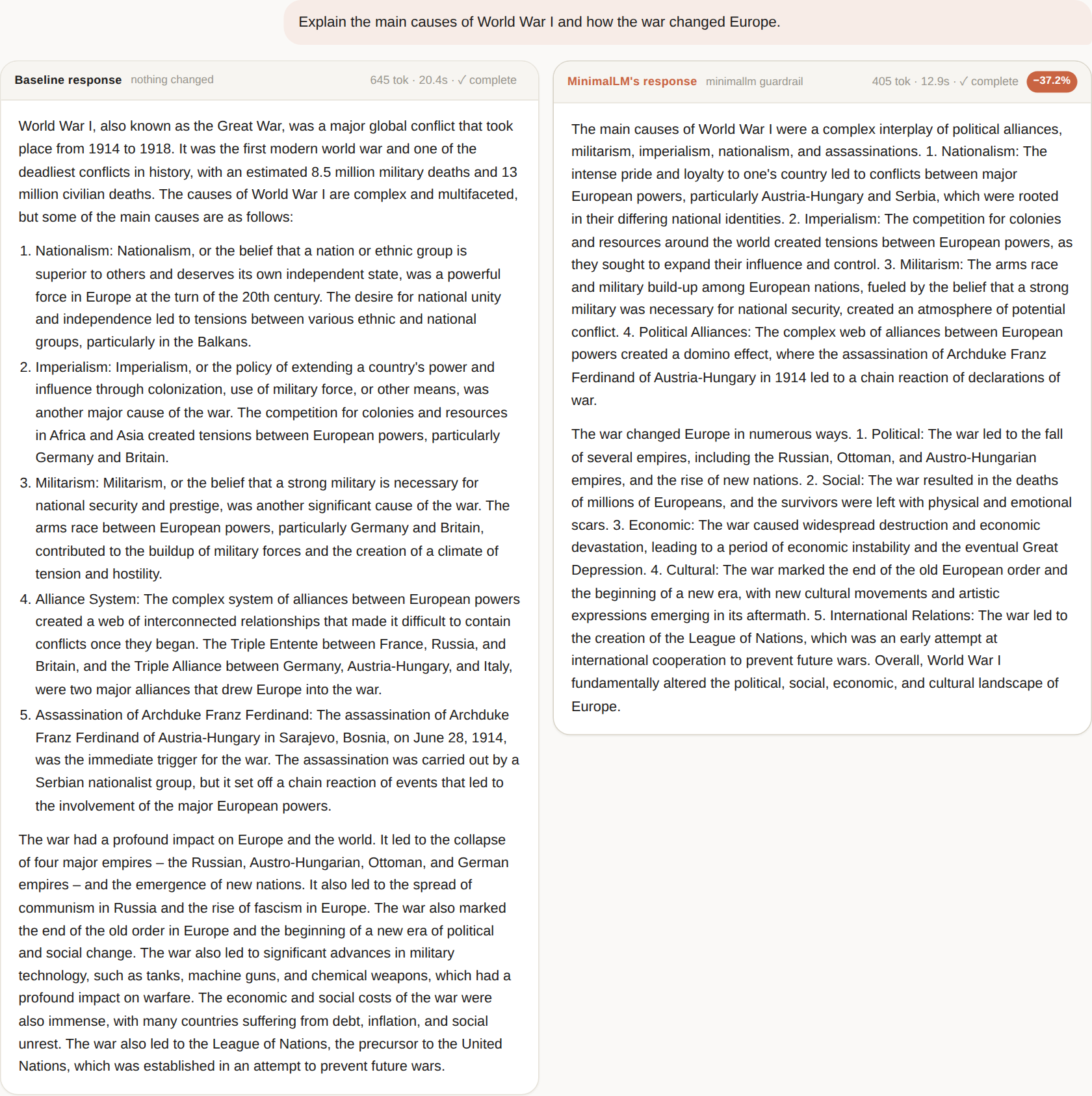}
\caption{Interface comparison on a history question (Mistral-7B). Left: baseline; right: MinimalLM.}
\label{fig:demo-history}
\end{figure*}

\begin{figure*}[p]
\centering
\includegraphics[width=\textwidth,height=0.82\textheight,keepaspectratio]{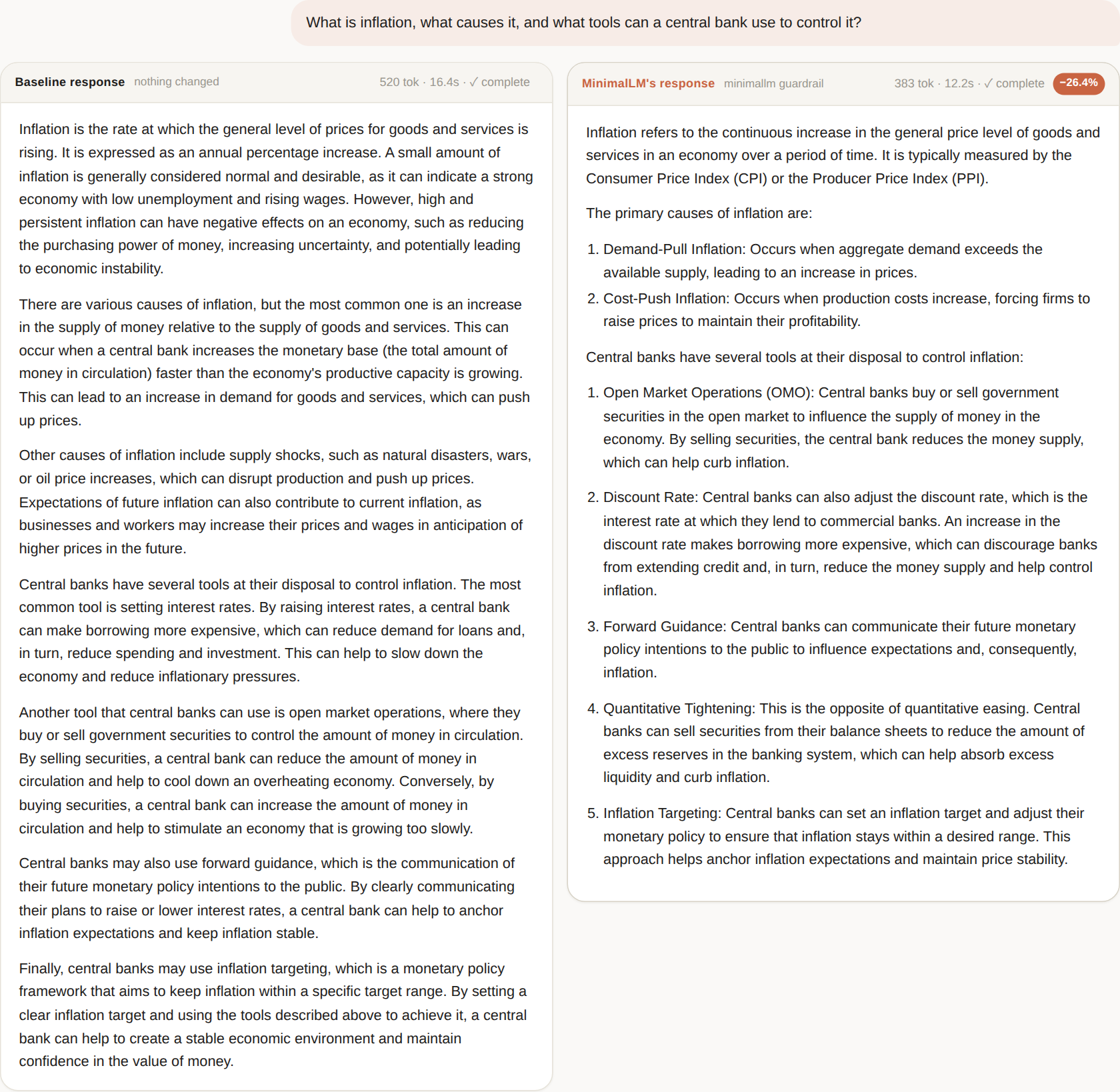}
\caption{Interface comparison on an economics question (Mistral-7B). Left: baseline; right: MinimalLM.}
\label{fig:demo-economics}
\end{figure*}

\begin{figure*}[p]
\centering
\includegraphics[width=\textwidth,height=0.82\textheight,keepaspectratio]{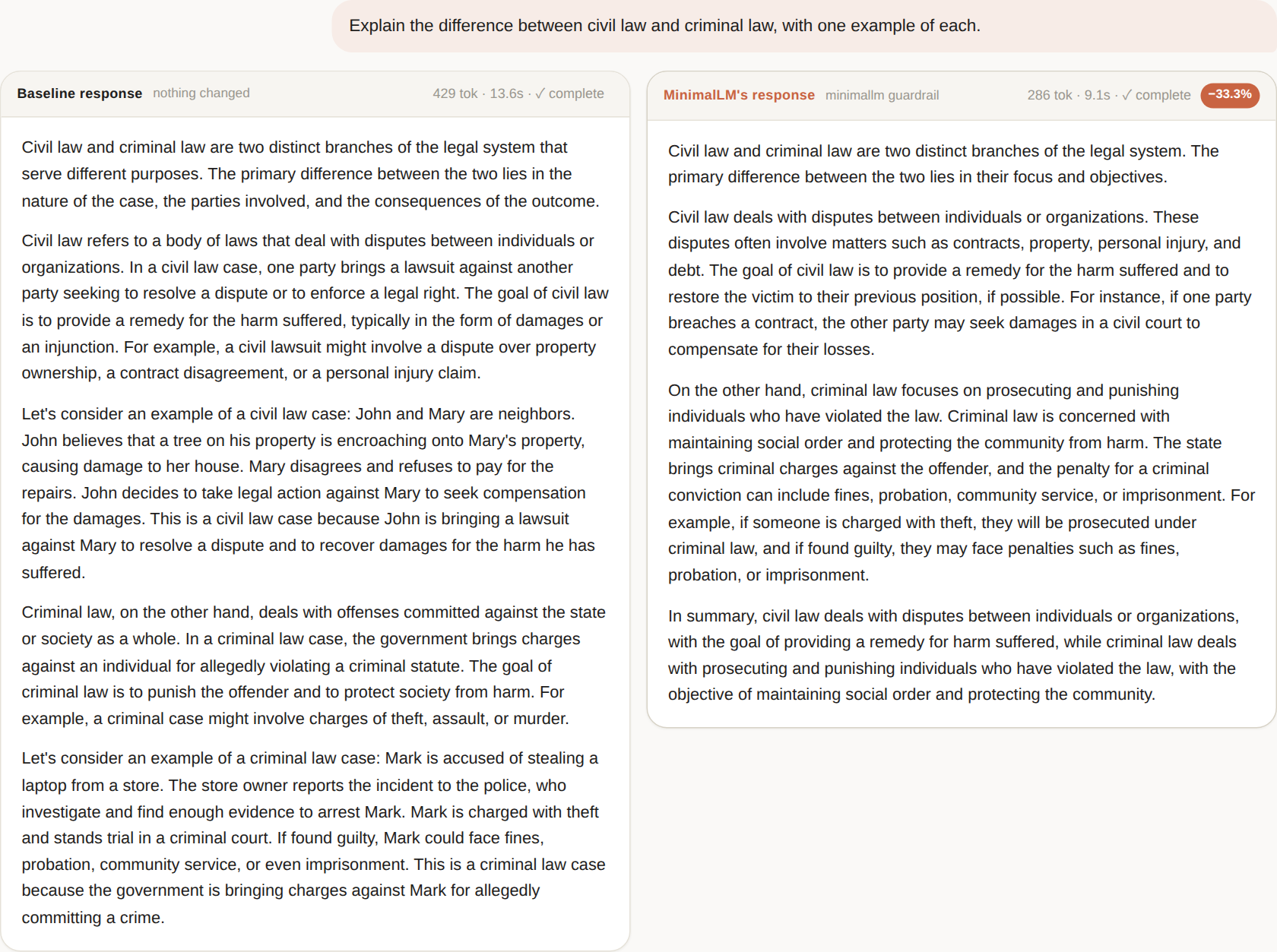}
\caption{Interface comparison on a law question (Mistral-7B). Left: baseline; right: MinimalLM.}
\label{fig:demo-law}
\end{figure*}

\begin{figure*}[p]
\centering
\includegraphics[width=\textwidth,height=0.82\textheight,keepaspectratio]{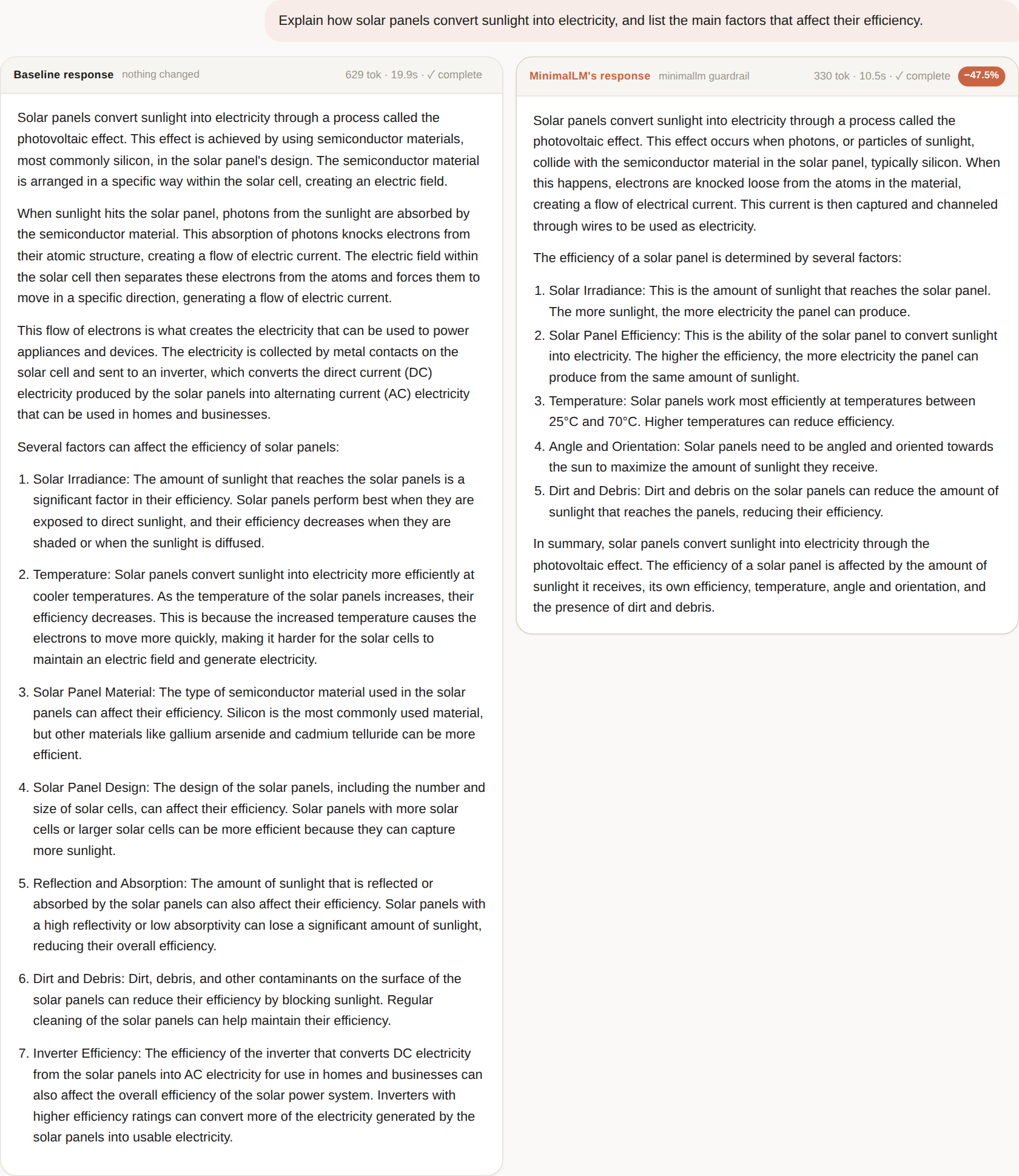}
\caption{Interface comparison on a science question (Mistral-7B). Left: baseline; right: MinimalLM.}
\label{fig:demo-science}
\end{figure*}

\begin{figure*}[p]
\centering
\includegraphics[width=\textwidth,height=0.82\textheight,keepaspectratio]{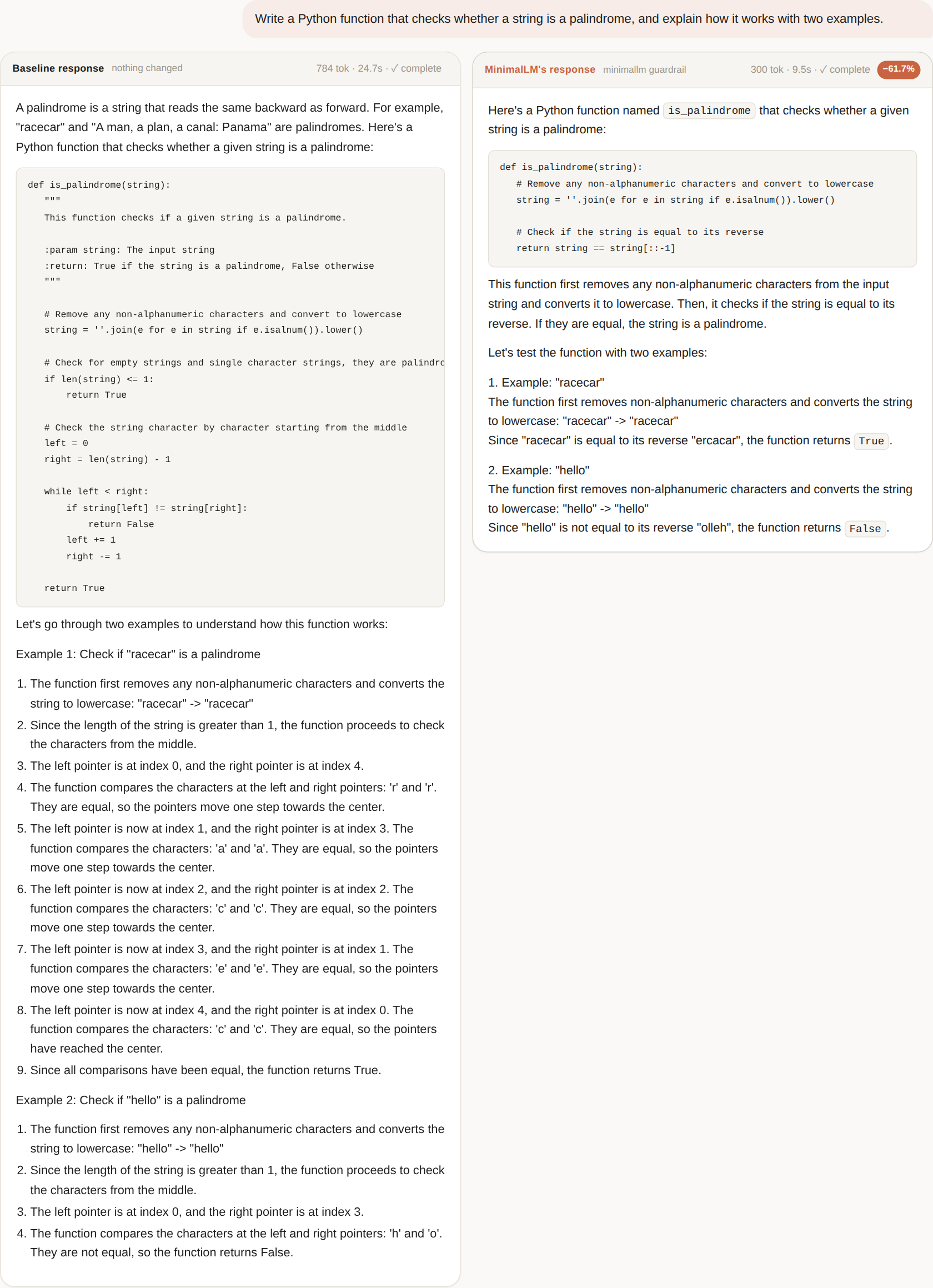}
\caption{Interface comparison on a programming question (Mistral-7B). Left: baseline; right: MinimalLM.}
\label{fig:demo-coding}
\end{figure*}

\begin{figure*}[p]
\centering
\includegraphics[width=\textwidth,height=0.82\textheight,keepaspectratio]{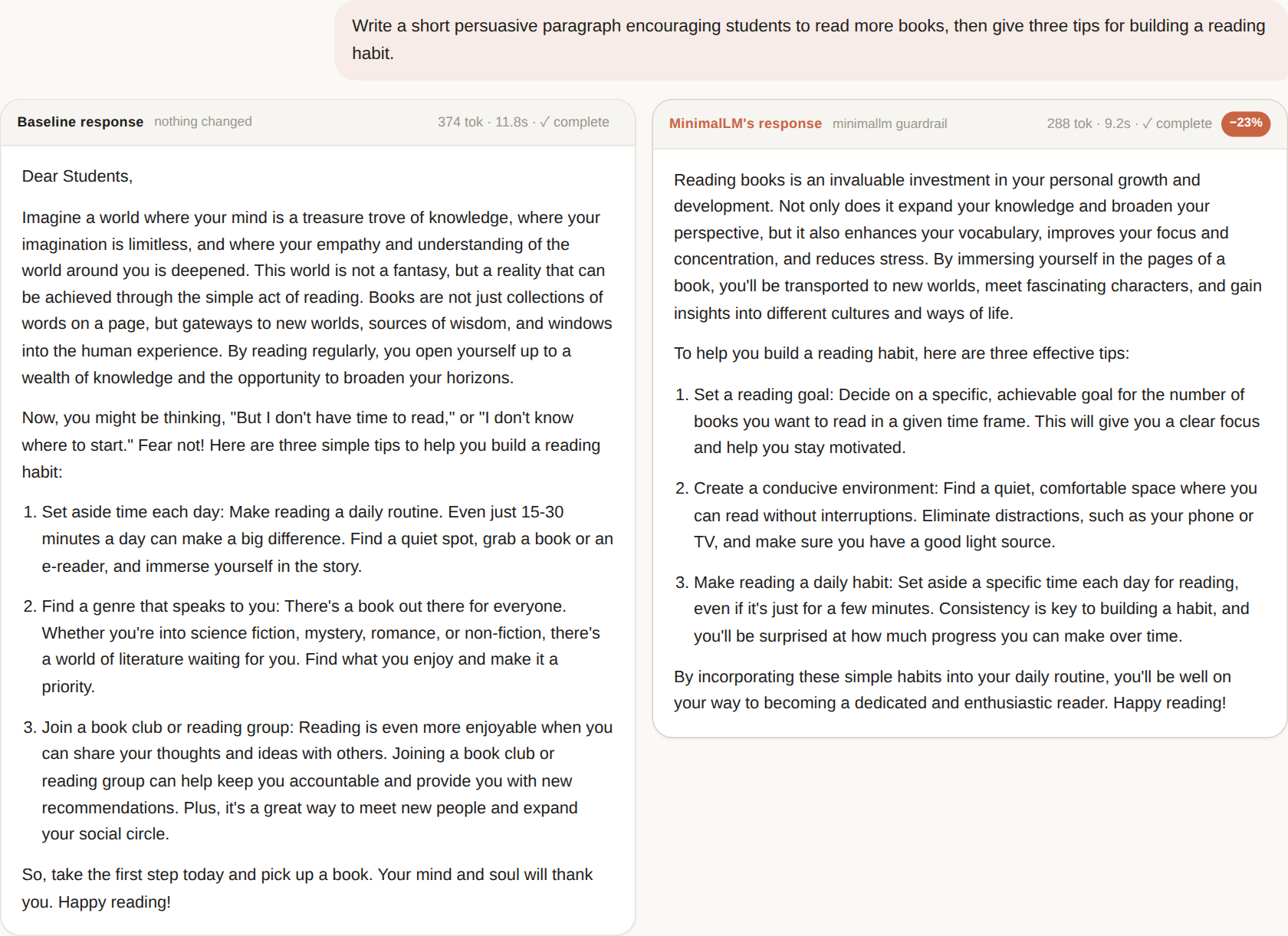}
\caption{Interface comparison on a writing question (Mistral-7B). Left: baseline; right: MinimalLM.}
\label{fig:demo-writing}
\end{figure*}

\begin{figure*}[p]
\centering
\includegraphics[width=\textwidth,height=0.82\textheight,keepaspectratio]{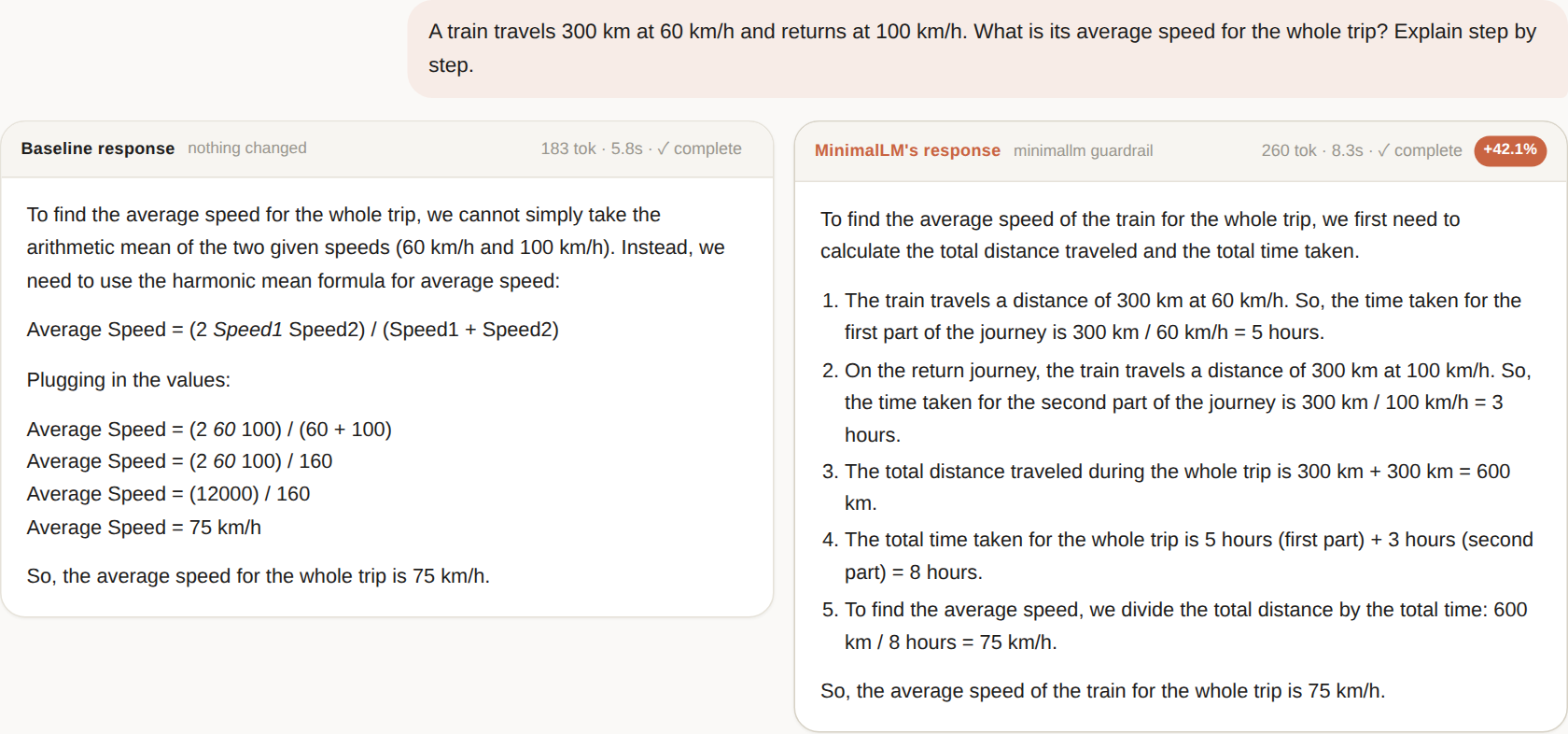}
\caption{Interface comparison on a mathematics question (Mistral-7B). Left: baseline; right: MinimalLM.}
\label{fig:demo-math}
\end{figure*}

\end{document}